# Vehicle classification based on convolutional networks applied to FM-CW radar signals

Samuele Capobianco, Luca Facheris, Fabrizio Cuccoli and Simone Marinai

**Abstract** This paper investigates the processing of Frequency Modulated-Continuos Wave (FM-CW) radar signals for vehicle classification. In the last years deep learning has gained interest in several scientific fields and signal processing is not one exception. In this work we address the recognition of the vehicle category using a Convolutional Neural Network (CNN) applied to range Doppler signature. The developed system first transforms the 1-dimensional signal into a 3-dimensional signal that is subsequently used as input to the CNN. When using the trained model to predict the vehicle category we obtain good performance.

## 1 Introduction

The automatic processing of real-time traffic information is an important feature to analyze the traffic. The number of traffic sensors is quickly growing as well as the amount of useful information available for traffic monitoring applications [1]. As we are moving towards Smart Cities, traffic monitoring becomes an important topic to address in order to improve the safety of traveling and to help the police monitoring work.

The previous mentioned peculiarities are required to build one intelligent transportation system where one important part is vehicle detection and classification. Vehicle classification is an important task also for police activities in order to prevent potential criminal behaviors.

---

Samuele Capobianco, Luca Facheris and Simone Marinai
Università degli studi di Firenze, via Santa Marta 3, Firenze
e-mail: {samuele.capobianco,luca.facheris,simone.marinai}@unifi.it

Fabrizio Cuccoli
CNIT RaSS c/o Dipartimento di Ingegneria dell'Informazione, via Santa Marta 3, Firenze
e-mail: fabrizio.cuccoli@unifi.it

Many solutions have been adopted to detect or classify vehicles based on different types of sensors ([2], [3], [4]). The traffic sensors produce a large quantity of data which can be useful as input to various machine learning techniques to address the vehicle detection and classification tasks. In particular, in this work we use deep learning to analyze signals coming from radars.

Deep learning addresses neural network architectures that are composed by several transformation layers which learn representations of the input data. These architectures compute multiple levels of abstraction which learn directly on the row data irrespective of application contexts. These methods have dramatically improved the state-of-the-art in speech recognition, visual object recognition, and many other application research fields.

One example of detection vehicles from camera videos is proposed for the automatic car counting. Using the OverFeat [5] framework (Convolutional Neural Network and SVM) and the Background Subtraction Method, the authors have demonstrated to be able to count cars. Another solution for a low-cost vehicle detection and classification system is based on a radar FM-CW [6]. The radar, unlike video sensors, is less vulnerable to weather and is able to work with the same performances in any light condition. In particular, the latter is a remarkable feature to build a reliable traffic control system.

In this paper, we address the vehicle classification in highway context by using information coming from a continuous wave radar. The Italian Traffic Law provides six categories of vehicles which can run on the highways. It is important to check the speed of running vehicles because each category has different speed limit. At present, the continuous wave radar is used by the major Italian highway company to monitor the speed of traveling motor vehicles on highway. We want to discover the right vehicle category using the radar sensor signal already employed to check the speed limit in order to help the police monitoring.

The rest of the paper is organizes as follows. In Section 2 we describe the main feature of the radar that provides the data used for vehicle classification. In Section 3 we analyze the overall system organization and the neural network architecture. The experiments performed are presented in Section 4, while concluding remarks are in the Conclusions (Section 5).

## 2 Signal Model of FM-CW Radar

In this work, we use signals collected by a 24 GHz Frequency-Modulated Continuos Wave radar in order to classify vehicles running on highways. As sketched in Figure 1, the radar was placed on a fixed structure at 5.3 m height with the antenna illuminating a single lane and pointing with a depression angle of $32^o$ toward the back of the vehicles passing under the structure along that lane.

As mentioned, data employed in this work come from a FM-CW radar. Like pulse Doppler radars, FM-CW radars can measure range and velocity of targets (vehicles in our case). However, the continuous signal transmission brings to the designer the benefit to reduce significantly the transmission power needed to transmit the same energy of a pulsed system, with a significant positive impact also on their size and cost. Furthermore, transmitter and receiver operate continuously, independently and simultaneously so that the receiver does not need to be silenced during the transmission. For the reasons briefly explained in the following, the price paid for these significant advantages is the need for a quite higher computational load with respect to pulsed radars. Nowadays, however, this is not a problem anymore: the availability of low cost, small size and high speed programmable FPGAs on the market is the reason why FM-CW radars are widely employed in a great number of applications, ranging from remote sensing to law enforcement and automotive industry. The waveform transmitted by a FM-CW radar is a frequency modulated signal:

$$A \cdot cos(2\pi f_0 t + 2\pi \int_{-\infty}^{t} m(\alpha) d\alpha) \qquad (1)$$

where the modulating signal $m(t)$ - coinciding with the instantaneous frequency deviation of the modulated signal - is a periodic triangular waveform with period $2T$

$$m(t) = \Delta f \cdot \sum_{n=-\infty}^{\infty} \left[ Tri\left( \frac{t - (n+1)T}{T} \right) - \frac{1}{2} \right] \qquad (2)$$

with

$$Tri(t) = \begin{cases} 1+t & \text{if } -1 < t \leq 0 \\ 1-t & \text{if } 0 < t < 1 \\ 0 & \text{elsewhere} \end{cases} \qquad (3)$$

and where $\Delta f$, that coincides in practice with the bandwidth of the modulated signal, is the variation of its instantaneous frequency during the so called "sweep interval" T. Note that, during two subsequent sweep intervals, the variation of the instantaneous frequency is of opposite sign: if it increases during the first interval (ascending or up-ramp) at a rate $\Delta f/T$, in the second interval (descending or down-ramp) it decreases of the same quantity and at the same rate. If the relative radial velocity between the radar and the target is zero, we have the situation sketched in Figure 2: the instantaneous frequency of the received (echo) signal (in green) will be the delayed replica of the transmitted signal (in red), the delay $\tau$ being equal to $2R/c$, where R is the range of the target and c the speed of light.

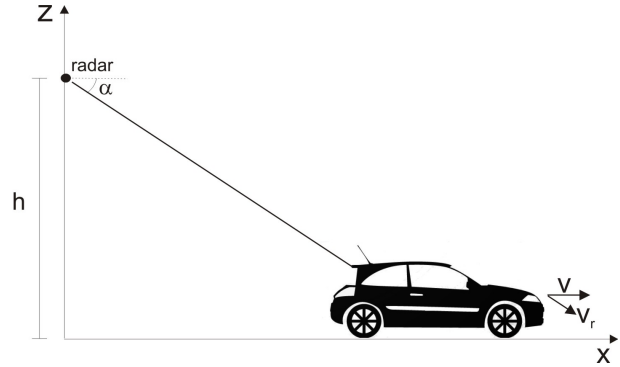

**Fig. 1** Scheme of the acquisition geometry of the FM-CW radar (h=5.3 m, $\alpha = 32^o$)

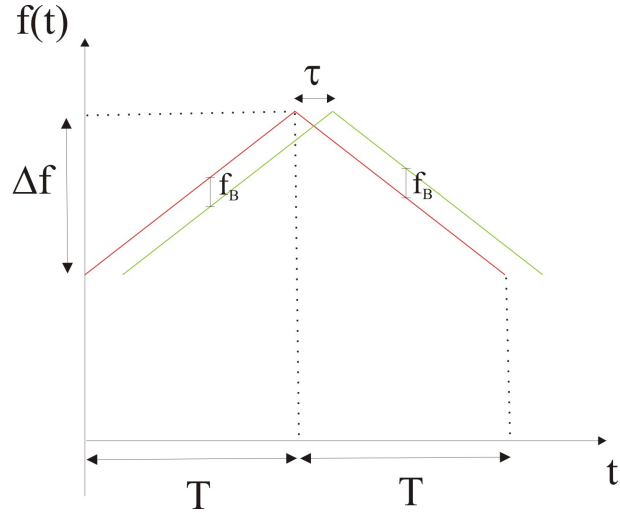

**Fig. 2** Instantaneous frequency of the transmit (red) and of the echo (green) signals for the case of absence of relative motion between the radar and the target. Only two ramps are plotted for the sake of clarity.

The received signal is converted to baseband by beating it with the reference transmit signal. This generates a sinusoidal signal with the same frequency $f_B$ (beat frequency) both in the up- and down-ramp case, and that is linearly related to the delay of the echo signal as follows

$$\tau = \frac{T}{\Delta f} \cdot f_B \qquad (4)$$

which allows to measure the range of the target. Therefore, in FM-CW systems the range scale corresponds to a frequency (beat) scale, while in pulsed systems it corresponds to a "simpler" time (delay) scale. This implies that a spectral analysis is needed in the FM-CW case, with the consequent higher computational load mentioned above. Consider in fact the case of several stationary targets: each of them will give rise to echo signals of different intensity that, beaten with the reference transmit signals, will generate sinusoidal contributions with different frequency at baseband. Intensive spectral (FFT) processing is therefore needed to detect targets and to resolve them in range. A further complication arises from the presence of a relative motion between radar and target. In such a case, assuming a relative radial velocity $v_r$, the received signal frequency is affected by a Doppler shift equal to $f_D = \pm \frac{2v_r}{\lambda}$, where $\lambda$ is the wavelength and the sign depends on the direction of the relative motion. In Figure 3 sketches the situation: the Doppler frequency is negative, so it could represent the case of Figure 1. In this case we get two different beat frequencies for the up- and down-ramps:

$$f_{B1} = \frac{\Delta f}{T} \cdot \tau + |f_D| \quad ; \quad f_{B2} = \frac{\Delta f}{T} \cdot \tau - |f_D| \tag{5}$$

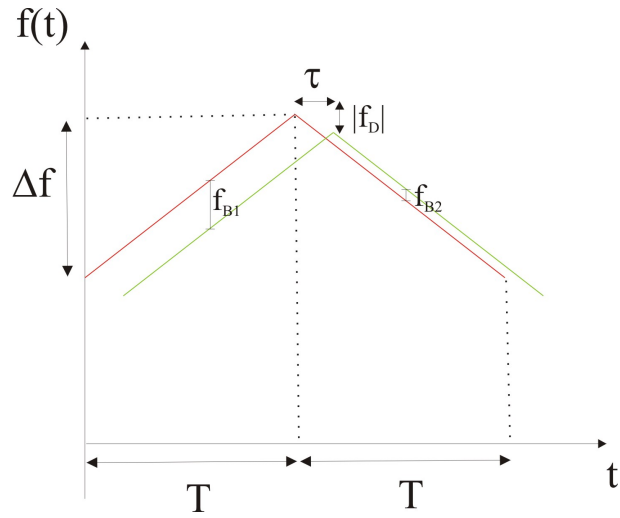

**Fig. 3** Instantaneous frequency of the transmit (red) and of the echo (green) signals for the case of presence of relative motion between the radar and the target. Only two ramps are plotted for the sake of clarity.

from which we can derive both delay and Doppler frequency of the target, and consequently its range and radial velocity. In the presence of multiple targets, determining their parameters with accuracy may ask for careful spectral processing.

In the case under exam, the interest is focused only on one target (the vehicle passing under the structure on which the radar is mounted). However, the objective is not only to determine its velocity, but also to classify the vehicle, which would be a challenge also for pulsed radars due to the number of parameters involved (speed, size, height, length, backscattering mechanisms involved and backscattered power) and to the possibility of artifacts (e.g. two different vehicles running close to each other). The 24 GHz radar parameters used for measurements are $\Delta f$=120 MHz and T=40 ms. It was decided to base the classification procedure on spectrograms obtained by 512 points FFT processing applied to the echo signals corresponding to the up- and down-ramps.

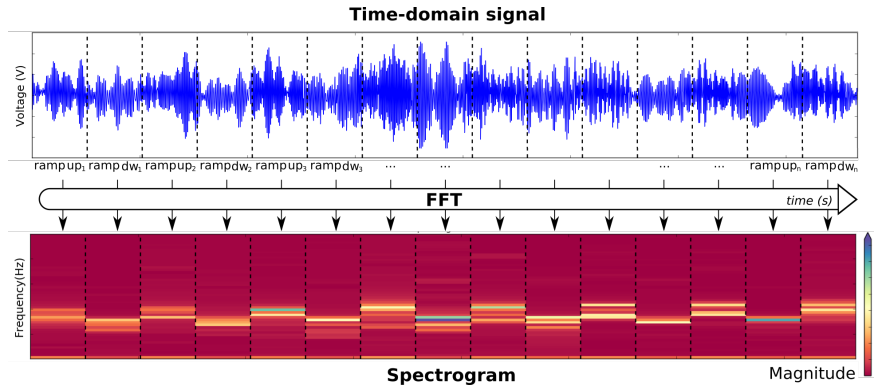

**Fig. 4** The received FM-CW radar signal is beaten with the transmitted signal: in the upper part of the figure is shown an example of the time domain signal obtained in that manner. The duration of the time slots (windows) shown in the upper picture is that of the ramps (up and down ramps) of the transmitted signals. Fast Fourier Transform (FFT) is then applied to each window. The two spectrograms that we utilized for further processing were derived by separating the upper ramp windows of the signal from the down ramp ones. The spectrogram corresponding to the up ramp was obtained by building a matrix whose columns are the FFT moduli of the sequence of up ramp windows. The spectrogram corresponding to the down ramp was obtained analogously. In the lower picture of the figure, the level of the computed modulus grows according to the color bar at the left of the lower picture.

## 3 Application scenario

Along the highway infrastructures are placed radar sensors used to monitor the vehicles speed by law enforcement to prevent potential criminal behaviors. In this work, we use the signal from FM-CW radar to classify the vehicles in order to replace the camera sensors typically used for this task.

### 3.1 Signal processing

The radar system used for monitoring the vehicles produces a beat frequency corresponding to each observed vehicle running along a lane. The signal that is then processed is a sequence of frames whose duration depends on the time taken by the vehicle to travel along the antenna footprint in the observed lane. As illustrated in Figure 5, the radar signal generated at baseband after having beaten the received signal with the transmitted one, is a sequence of received signals corresponding to the alternate sequences of transmitted up- and down-ramps.

One of the typical ways to analyze the radar beat frequency through time is to compute the so called range-Doppler signature. Such an approach has been already used to detect vehicles in a similar context [6]. Using the Short Time Fourier Transform (STFT) we are able to carry out the time-frequency analysis. STFT is a well known technique in signal processing to analyze non-stationary signals in the time domain. Given a non-stationary signal, a moving window of fixed duration selects a sequence of signal segments and computes a sequence of Fast Fourier Transform (FFT) on them. Windows may overlap or not for reasons related to optimal spectrum estimation. This procedure allows to analyze the spectral properties of the non stationary signal during their variations over time. Figure 4 depicts the process we used to compute the spectrograms from the original signals. Considering one of our FM-CW radar signals and using a moving window (without overlap) with a duration equal to the ramp duration (T), we extracted two sequences of alternate signal segments. For each segment we computed the FFT modulus that was then used to build the spectrogram, the first one corresponding to the sequence of up-ramp signals, the second one to that of down-ramp signals. Figure 5 shows the baseband signal collected from a vehicle and the two spectrograms obtained by applying the STFT algorithm. These two representations describe each vehicle running along the observed lane.

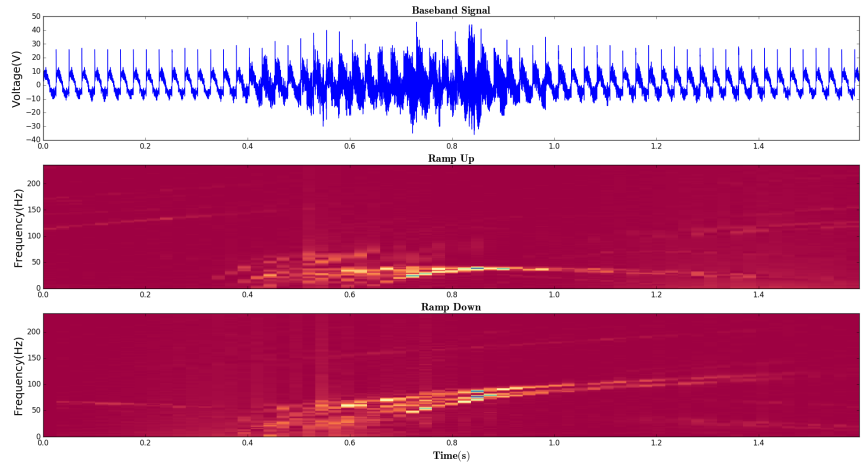

**Fig. 5** Radar signals. Top frame: baseband signal obtained by beating the received signal with the transmitted signal generated by one running vehicle observed by a FM-CW radar. In the middle and in the bottom frames are respectively shown the range-Doppler signature for the up- and down-ramp.

The idea to classify the running vehicles on the highway is to learn directly from the range-Doppler signature inherent to each spectrogram a data representation useful to discriminate the categories. In this case, the time-frequency analysis transforms a 1-dimensional signal (top signal in Figure 5) into 2-dimensional signals (up-ramp and down-ramp in Figure 5). Since we have a representation similar to one image, it is possible to use a neural network based on Convolutional Neural Network architecture to capture the topological structure of the input spectrograms and use the learned representation to classify the signals.

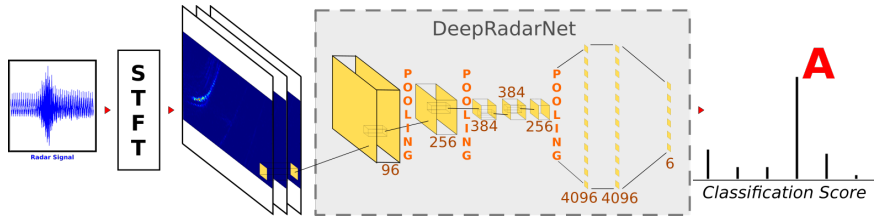

**Fig. 6** Given a 1-dimensional radar signal, using the Short Time Fourier Transformation we are able to compute three range Doppler signatures (up-ramp, down-ramp, average-ramp). Then, these three computed channels are stacked into one tensor. A trained *DeepRadarNet* model maps the tensor into the classification score predicting the class of vehicle.

## *3.2 DeepRadarNet*

In the last years, Convolutional Neural Networks obtained very good results in several different tasks. These architectures are used for Object Recognition tasks ([7], [8], [9]) improving also the accuracy score with respect to human performance [10]. These architectures require very large datasets to learn all their parameters. For instance, one of the most popular datasets used to train these networks is the Imagenet [11] one that contains 3.2 millions of images according to the WordNet hierarchy.

Convolutional Neural Networks are a particular type of artificial neural network consisting of alternated convolutional and spatial pooling layers [12]. The convolutional layers generate feature maps by linear convolutional filters followed by non-linear activation functions (e.g. rectifier, sigmoid, tanh). In 2012 the AlexNet [13] Convolutional Neural Network outperformed other machine learning techniques which were based on hand-crafted features in the object recognition contest on the Imagenet dataset.

Our proposed architecture, named *DeepRadarNet*, is inspired to the AlexNet architecture. We depict the model architecture in Figure 6.

The model consists of eight main layers (five convolutional and three fully connected layers). The first convolutional layer has 96 kernels with a receptive field of $3 \times 11 \times 11$ size and a stride of 4 units. The output of the first layer is then pooled to the first max-pooling layer where the max pooling and response normalization has been performed. The second layer has 256 kernels with a receptive field of $96 \times 5 \times 5$ size that is followed by another max-pooling layer. The max-pooling computes a max operator with kernel $3 \times 3$ and stride $2 \times 2$. The third, fourth and fifth convolutional layers have a different number of kernels and receptive field dimensions. Third layer: 384 kernels $256 \times 3 \times 3$ receptive field; fourth layer: 384 kernels and $384 \times 3 \times 3$ receptive field: fifth layer 256 kernels and $384 \times 3 \times 3$ receptive fields. The last convolution layer is followed by the last max-pooling. The computed features are the input for three fully connected layers which map the computed features into the output neurons. The sixth and the seventh layer compute a fully-connected transformation and each layer is composed by 4096 neurons. To avoid overfitting, the dropout [14] layer is applied between the previous fully-connected layers. The eighth layer corresponds to the output neurons that match the number of classes.

The output of the last fully-connected layer is fed into an N-way softmax which models a distribution over the $N$ class labels. Learning the model parameters is an optimization task which can be solved using multi class cross-entropy cost function used during the training phase.

In this work we consider a classification task with six vehicle categories where each category is associated to an output neuron. Using this architecture we can classify the input signal into vehicle categories according to the Italian Traffic Law.

### *3.3 Overall system structure*

Given a signal from the FM-CW radar, we can compute two spectrograms which describe the running vehicle on an observed segment of highway lane. After the STFT transformation, we have two spectrograms (2-dimensional signals) as range Doppler signature. The length of these signals depends on the vehicle speed. In order to have the same dimension for each observation and feed these data into the neural network we need to uniform the input shape. To this purpose, we fix the input shape according to the sample size distribution. Through zero-padding, we adjust the computed spectrograms obtaining a tensor composed by three channels, one for the up-ramp, one for the down-ramp and the last one for the average between the previous channels. We therefore have one tensor with a fixed size for each sample.

The three channels obtained in this manner are useful as input to the proposed neural network, which maps one sample to the classification score that correspond to the category of the observed vehicle. The input tensors are mean normalized using the train set average tensor. The mean compensation is applied both on the train and test set signal tensor.

After the training phase, we use the best trained model to predict the vehicle category. As shown in Figure 6, the input signal goes through the STFT operator which computes three 3-dimension tensors: one channel for each signal representation (up-ramp, down-ramp, average-ramp). After the zero-padding size normalization, the category is predicted from the trained *DeepRadarNet* model. The classification score is a way to discriminate the vehicle category.

## 4 Experiments

In this section we describe the experiments carried out to evaluate the proposed solution. As detailed in Figure 7 the dataset is composed by $9,981$ 1-dimensional signals not uniformly distributed in the six categories: car (A), car-trailer (B), truck (C), cargo truck (D), bus (E) and motorcycle (G).

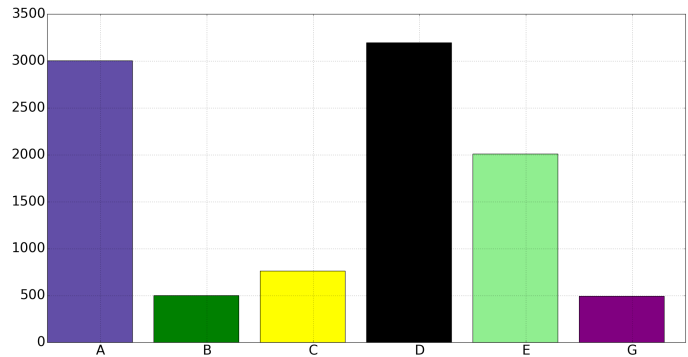

**Fig. 7** Distribution of the 9,981 samples into the 6 classes. We can see as the classes are not uniformly distributed. In particular, we have more examples for car, cargo truck and bus categories (A, D and E classes)

As we have seen in Section 3.1 it is possible to compute the range Doppler signature and visualize the spectrograms as show in Figure 8.

Each signal length depends on the vehicle speed during the observation since the number of frames for each spectrogram is proportional to the vehicle speed. We therefore need to normalize this input size to a fixed size in order to use the signals as input to the model. In Figure 7 is depicted the sample distribution.

## 4.1 10-fold cross-validation

To evaluate the system capability to discriminate the vehicle categories from the range Doppler signature we use the *k*-fold cross validation. In this case we extract randomly 10 different folds each composed by 2400 samples for training, 270 samples for validation and 7311 samples for testing. In particular, the train and validation set have 400 samples and 45 samples for each category respectively. We shuffled the dataset before extracting each fold.

During the training phase we use the stochastic gradient descent to adjust the model parameters. We use one batch size with one sample for each category (six samples for each batch). The learning rate, momentum, and weight decay are set to 0.0001, 0.9 and 0.0005 respectively.

One of the problems for training deep neural networks is the parameters initialization, especially when the training set is not large enough. One possible solution is to transfer the learned weights from another application context [15]. In our case we initialize the convolution layer parameters using the Alexnet pre-trained model [13] exception made for the last fully connected layers, that are randomly initialized.

In Figure 9 we depict the confusion matrix for one folder and the average results for all the folders. We have obtained good results considering that this one preliminary experiment. Having a look to results the class G (motorcycle) obtains a prediction accuracy of 1 which corresponds to a perfect classification score.

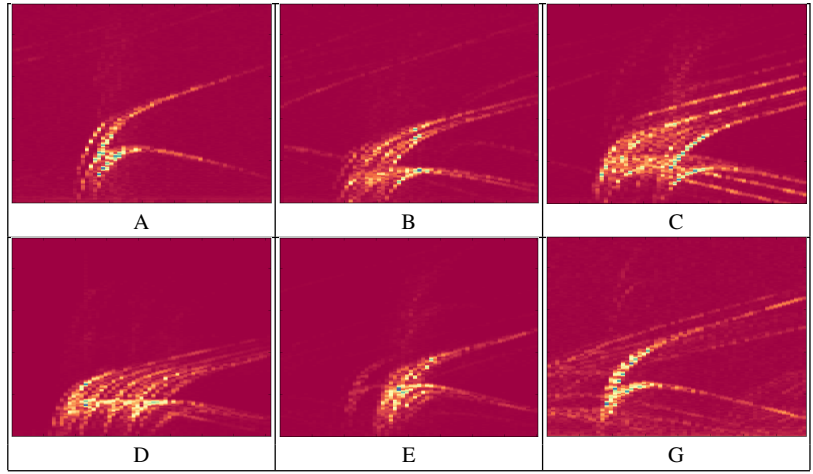

**Fig. 8** One transformed example for each category. We represent the range-Doppler signature as the average signal between the up- and the down-ramp. The meaning of the letters is the following: A: car, B: car-trailer, C: truck , D: cargo truck, E: bus, G: motorcycle.

For the classes A, B, and D ,which are car, car-trailer and cargo truck categories respectively, we obtain good results may be due to the network is able to describe the shape related to their spectrograms. For these three classes we can see in Figure 8 some examples from the dataset and we can get as these spectrograms have discriminative shapes though a large portion of them are quite similar to each other.

For the classes C and E, which are truck and bus respectively, we have some misclassification due to the related spectrogram shapes. Always in Figure 8, if we consider the vehicle structure (captured by the radar signal), the categories C and E include different shapes which are more difficult to discriminant and classify.

## 5 Conclusions

In this preliminary work we wanted to investigate the capability of Convolutional Neural Networks to recognize vehicle categories analyzing a FM-CW radar signal. In particular, we have proposed one technique to transform a 1-dimensional signal to a 3-dimensional tensor based on Short Time Fourier Transformation. The computed tensors are then used to train our convolutional architecture named *DeepRadarNet*.

Using transfer learning technique it is possible to initialize the model weights moving the learned parameters from a pre-trained model from Object Recognition task to our *DeepRadarNet* model. In this way the training phase starts from a better starting point improving also the recognition performances. We obtained good results in this prediction task which encourage us to continue the research in this way.

In the future work we want to address also the vehicle detection directly on the radar signal and improve the performance in this vehicle classification task. In particular, we want to compare different deep learning architectures considering also Recurrent Neural Networks. The idea could be to compare these different models with performance measures and also memory consumption in order to develop a stand alone low-cost system able to work on cheaper single-board computer.

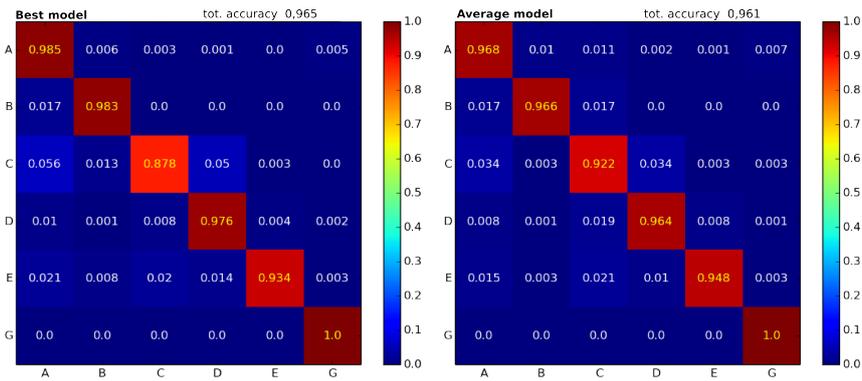

**Fig. 9** The class confusion matrix of the results obtained. The average recognition accuracy is 0.961 considering ten different folds. On the left the confusion matrix for one of the best folders. On the right the average results considering all folders.

**Acknowledgements** The authors wish to tank Infomobility S.R.L. Concordia sulla Secchia (Modena) and Autostrade per l'Italia (Roma) for having provided the radar data.

# References

1. J. M. Munoz-Ferreras, J. Calvo-Gallego, and F. Perez-Martinez. Monitoring road traffic with a high resolution lfmcw radar. In *IEEE Radar Conference*, pages 1–5, May 2008.
2. Y. K. Ki and D. K. Baik. Vehicle-classification algorithm for single-loop detectors using neural networks. *IEEE Transactions on Vehicular Technology*, 55(6):1704–1711, 2006.
3. G. De Angelis, A. De Angelis, V. Pasku, A. Moschitta, and P. Carbone. A simple magnetic signature vehicles detection and classification system for smart cities. In *IEEE International Symposium on Systems Engineering (ISSE 2016)*, pages 1–6.
4. H. Sandhawalia, J. A. Rodriguez-Serrano, H. Poirier, and G. Csurka. Vehicle type classification from laser scanner profiles: A benchmark of feature descriptors. In *16th International IEEE Conference on Intelligent Transportation Systems (ITSC 2013)*, pages 517–522.

5. Pierre Sermanet, David Eigen, Xiang Zhang, Michaël Mathieu, Robert Fergus, and Yann Lecun. *Overfeat: Integrated recognition, localization and detection using convolutional networks*. 2014.
6. J. Fang, H. Meng, H. Zhang, and X. Wang. A low-cost vehicle detection and classification system based on unmodulated continuous-wave radar. In *2007 IEEE Intelligent Transportation Systems Conference*, pages 715–720, Sept 2007.
7. Christian Szegedy, Wei Liu, Yangqing Jia, Pierre Sermanet, Scott Reed, Dragomir Anguelov, Dumitru Erhan, Vincent Vanhoucke, and Andrew Rabinovich. Going deeper with convolutions. In *CVPR 2015*, 2015.
8. K. Simonyan and A. Zisserman. Very deep convolutional networks for large-scale image recognition. *CoRR*, abs/1409.1556, 2014.
9. Jonathan Long, Evan Shelhamer, and Trevor Darrell. Fully convolutional networks for semantic segmentation. *CVPR (to appear)*, November 2015.
10. Kaiming He, Xiangyu Zhang, Shaoqing Ren, and Jian Sun. Delving deep into rectifiers: Surpassing human-level performance on imagenet classification. In *Proceedings of the IEEE International Conference on Computer Vision (ICCV 2015)*, pages 1026–1034.
11. J. Deng, W. Dong, R. Socher, L.-J. Li, K. Li, and L. Fei-Fei. ImageNet: A Large-Scale Hierarchical Image Database. In *CVPR09*, 2009.
12. Yann Lecun, Léon Bottou, Yoshua Bengio, and Patrick Haffner. Gradient-based learning applied to document recognition. In *Proceedings of the IEEE*, pages 2278–2324, 1998.
13. Alex Krizhevsky, Ilya Sutskever, and Geoffrey E. Hinton. Imagenet classification with deep convolutional neural networks. In F. Pereira, C.J.C. Burges, L. Bottou, and K.Q. Weinberger, editors, *Advances in Neural Information Processing Systems 25*, pages 1097–1105. Curran Associates, Inc., 2012.
14. Nitish Srivastava, Geoffrey Hinton, Alex Krizhevsky, Ilya Sutskever, and Ruslan Salakhutdinov. Dropout: A simple way to prevent neural networks from overfitting. *Journal of Machine Learning Research*, 15:1929–1958, 2014.
15. Jason Yosinski, Jeff Clune, Yoshua Bengio, and Hod Lipson. How transferable are features in deep neural networks? In Z. Ghahramani, M. Welling, C. Cortes, N.d. Lawrence, and K.q. Weinberger, editors, *Advances in Neural Information Processing Systems 27*, pages 3320–3328. Curran Associates, Inc., 2014.